\documentclass{article}
\usepackage{spconf,amsmath,graphicx}
\usepackage[justification=centering]{caption}
\usepackage{booktabs,dsfont,enumitem,hyperref}
\setlist[description]{labelindent=1cm,leftmargin=3cm,style=multiline}

\title{ROBUSTNESS AND OVERFITTING BEHAVIOR OF IMPLICIT BACKGROUND MODELS}
\name{Shirley Liu, Charles Lehman, and Ghassan AlRegib}
\address{OLIVES at the Center for Signal and Information Processing (CSIP),\\
School of Electrical and Computer Engineering,\\
Georgia Institute of Technology, Atlanta, GA, 30332-0250 USA\\
\{shirley.liu, charlie.k.lehman, alregib\}@gatech.edu}

\begin{document}
\onecolumn 

\begin{description}

\item[\textbf{Citation}]{S. Liu, C. Lehman and G. AlRegib, "Robustness and Overfitting Behavior of Implicit Background Models," IEEE International Conference on Image Processing (ICIP), Abu Dhabi, United Arab Emirates, Oct. 2020.
} \\



\item[\textbf{Code}]{\url{https://github.com/olivesgatech/scribe-robustness-overfit}} \\

\item[\textbf{Bib}] {
@INPROCEEDINGS\{Liu2020,\\ 
author=\{S. Liu, C. Lehman and G. AlRegib\},\\ 
booktitle=\{IEEE International Conference on Image Processing (ICIP)\},\\ 
title=\{Robustness and Overfitting Behavior of Implicit Background Models\},\\ 
year=\{2020\}\}\\
} \\

\item[\textbf{Copyright}]{\textcopyright 2020 IEEE. Personal use of this material is permitted. Permission from IEEE must be obtained for all other uses, in any current or future media, including reprinting/republishing this material for advertising or promotional purposes,
creating new collective works, for resale or redistribution to servers or lists, or reuse of any copyrighted component
of this work in other works. } \\

\item[\textbf{Contact}]{\href{mailto:alregib@gatech.edu}{alregib@gatech.edu}~~~~~~~\url{https://ghassanalregib.info/} \\ \href{mailto:charlie.k.lehman@gmail.com
}{charlie.k.lehman@gmail.com~~~~~~~\url{https://charlielehman.github.io/}
} \\ \href{mailto:shirley.liu@gatech.edu}{shirley.liu@gatech.edu}}
\end{description} 

\thispagestyle{empty}
\newpage
\clearpage

\twocolumn

\ninept
\maketitle

\begin{abstract}
\vspace{-0.5em}

In this paper, we examine the overfitting behavior of image classification models modified with Implicit Background Estimation (SCrIBE), which transforms them into weakly supervised segmentation models that provide spatial domain visualizations without affecting performance. 
Using the segmentation masks, we derive an overfit detection criterion that does not require testing labels. 
In addition, we assess the change in model performance, calibration, and segmentation masks after applying data augmentations as overfitting reduction measures and testing on various types of distorted images.

\end{abstract}

\vspace{-0.5em}
\section{Introduction}
\label{sec:intro}
\vspace{-0.5em}

As real-world applications of deep learning models become more commonplace, so does the need to provide a means of intuitive feedback to users and domain experts regarding the reliability of outputs in a variety of scenarios.
In the case of a vehicle mounted camera, where the inputs are natural images, a model may encounter inputs that are corrupted due to environmental effects such as weather, motion blur, or lighting.
When that corrupted image is part of a critical decision system, it is imperative that any failures of the vision system are handled safely.
This means that the outputs from a corrupted input are either blocked from contributing or the model performs nominally despite the corruption.
Though the latter is more desirable, it is very difficult to prove to hold outside of controlled experimentation in a lab setting. 
Another challenge we face is that labeled training data may not be readily available in real-world settings, resulting in the risk of overfitting models. Oftentimes it is difficult to determine whether a model overfits and its degree of overfit. 
Our goal here is to explore a means of providing feedback to a user regarding the behavior of image classifiers under \emph{corrupted inputs} and \emph{limited training data}.

\begin{figure}[h!]
  \centering
  \includegraphics[width=1\linewidth]{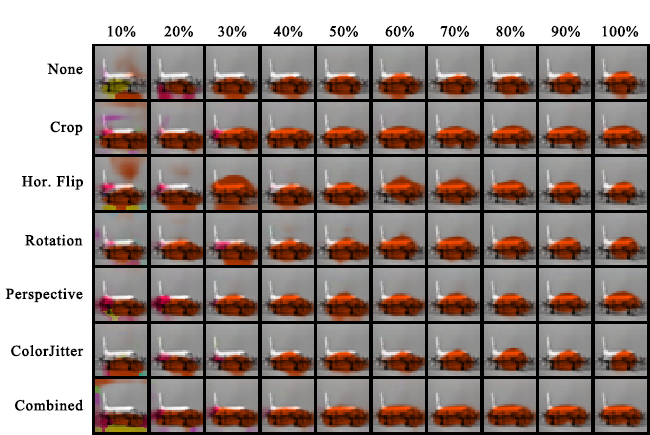}
  \caption{Segmentation masks by subset size and data augmentation for a sample plane image.}
  \label{fig:subset_aug_seg}
  \vspace{-1em}
\end{figure}

In an image classification model, numerical measures on the model performance are available, but we can acquire more insight if visual information about what the model inspects when making a classification decision are given on a pixel-wise level. 
Lehman et al. propose a concept called Implicit Background Estimation (IBE) in \cite{ibe} that expresses the background in an image in terms of the foreground and employs it in semantic segmentation tasks. 
IBE can be incorporated into image classification tasks as well by making minor changes to the neural network structure and training criterion. 
In this way, the classification model is converted into a weakly supervised segmentation model that provides coarse segmentation masks using only image-level labels. 
Through these visualizations in the spatial domain, we can better study the effects of overfitting and input perturbation on the models and how different data augmentations can alter their performance and calibration under each circumstance.

Several intrinsic and extrinsic methods for detecting model overfit have been presented in literature. 
Intrinsic methods do not require using a validation set in their detection; they only depend on the model and training data. These include detection via adversarial examples \cite{overfit_adversarial} 
and counterfactual simulation \cite{overfit_circuit}. 
Extrinsic methods utilize testing data in determining overfit, such as the traditional method of comparing testing and training fitness \cite{overfit_trad} and inspecting the correlation of fitness values on training and testing sets \cite{overfit_corr}. 
A common drawback of current overfit detection measures is the dependence on testing labels and metrics related to accuracy or loss. 
To the best of our knowledge, the only method that uses prediction maps as visual output to detect overfit is considerably qualitative, and the quantitative metric it proposes requires the use of testing labels \cite{overfit_spec}. 
In contrast, in this paper we present a quantitative metric for overfit detection attained through the segmentation masks produced by a SCrIBE model that does not require testing labels.

\begin{table*}[t!]
\centering
\scalebox{0.9}{
\begin{tabular}{ccccccccccc}
\toprule
& \textbf{10\%} & \textbf{20\%} & \textbf{30\%} & \textbf{40\%} & \textbf{50\%} & \textbf{60\%} & \textbf{70\%} & \textbf{80\%} & \textbf{90\%} & \textbf{100\%} \\
\midrule[\heavyrulewidth]
\textbf{No Transform} & 0.762 & 0.801 & 0.824 & 0.820 & 0.829 & 0.834 & 0.846 & 0.858 & 0.867 & 0.872 \\
\midrule
\textbf{RandomCrop} & 0.495 & 0.669 & 0.722 & 0.752 & 0.765 & 0.762 & 0.779 & 0.786 & 0.792 & 0.801 \\
\midrule
\textbf{RandomHorizontalFlip} & 0.664 & 0.748 & 0.773 & 0.808 & 0.813 & 0.821 & 0.833 & 0.841 & 0.851 & 0.861 \\
\midrule
\textbf{RandomRotation} & 0.706 & 0.792 & 0.811 & 0.817 & 0.834 & 0.844 & 0.849 & 0.854 & 0.864 & 0.869 \\
\midrule
\textbf{RandomPerspective} & 0.603 & 0.720 & 0.770 & 0.786 & 0.796 & 0.804 & 0.819 & 0.825 & 0.840 & 0.846 \\
\midrule
\textbf{ColorJitter} & 0.707 & 0.814 & 0.828 & 0.832 & 0.831 & 0.845 & 0.854 & 0.864 & 0.864 & 0.872 \\
\midrule
\textbf{Combined} & 0.439 & 0.536 & 0.607 & 0.654 & 0.677 & 0.699 & 0.713 & 0.727 & 0.751 & 0.760 \\
\bottomrule
\end{tabular}
}
\caption{Sparsity measures for different subset size and data augmentation training configurations.}
\label{table:sparsity}
\vspace{-0.5em}
\end{table*}

Data augmentation, which increases the size of labeled training data, has proven to be successful in alleviating overfitting and is a heavily studied area. Several creative data augmentation methods have been developed 
\cite{da_effectiveness, da_adaptive, da_implicit, da_cmp_2, da_bayesian}. 
Some of these methods, such as random erase \cite{da_randerase} and adversarial data augmentation \cite{da_adversarial}, are robust to occlusions or unseen data distributions. Several experiments investigate the types and combinations of traditional transform data augmentation in relation to their success in improving the performance of overfit models. \cite{da_exp} and \cite{da_exp_2} reach the similar conclusion that data augmentation is especially helpful for models trained with a very small training set, and that a few simple augmentations are more efficient than layering many complex ones. \cite{da_exp_2}, \cite{da_cmp}, and \cite{da_cmp_2} further conclude that geometric transforms outperform color transforms, with cropping, flipping, and rotation being the most effective. \cite{da_cmp_2} demonstrates that cropping provides the largest individual improvement, cropping with flipping is the best combination, and color transforms work well when used in conjunction with cropping and flipping. To the best of our knowledge, there has not been experiment performed that assesses the robustness of data-augmented models with perturbed testing data, nor any that explores the relationship between data augmentations and a model's degree of overfit and robustness. Experiments mentioned above that do examine degrees of overfit and data augmentation draw conclusions based on quantitative measures like accuracy, but do not provide an intuition as to how different data augmentation affect the model's perception. These are the major topics we will explore.

In this paper, we will \emph{demonstrate that the segmentation masks obtained through SCrIBE models reveal the model's level of overfitting, and an overfit detection metric can be derived from them without the need of testing image labels}. The performance and calibration of models trained using various numbers of training samples and several data augmentations are assessed on both the original testing set and the testing sets distorted with different types and levels of perturbation. We show that the segmentation masks provide meaningful intuition on the distortions present in the input images as well as the calibration of the model.
\vspace{-0.5em}
\section{Method}
\label{sec:method}
\vspace{-0.5em}
\subsection{Weak Segmentation with Implicit Background}
\vspace{-0.5em}
\label{ssec:weak_seg}

By combining Implicit Background Estimation (IBE) \cite{ibe} and LogSumExp Pooling (LSE) \cite{pinheiro2015image}, we propose \emph{SCrIBE}, a simple method for learning segmentations predictions while training with only image-level labels.
LSE enables a tunable spatial aggregation to transition between a Class Activation Map (CAM) and a logit vector to enable optimization with a one-hot vector.
However, \cite{pinheiro2015image} uses background images during training in addition to several post-processing techniques.
As background is the complement of the foreground, it is natural to represent it as such in optimization through IBE.
By replacing Softmax with Sigmoid, $\sigma$, we can update through each component $n$ at pixel location $i,j$ for logit, $\mathbf{v}_{i,j}$ and pixel label $y_{i,j}$ in the CAM by the following gradient update:
\begin{align*}
    \frac{\delta L}{\delta v_{i,j,n}} =  \begin{cases}
    -\frac{e^{v_{i,j,n}}}{1+\sum_{\text{FG}}e^{v_{i,j,m}}} \qquad\qquad n = \text{Background}\\
    -y_{i,j,n}(1-\sigma(v_{i,j,n})) \qquad n \in \text{Foreground}\\
    \end{cases}
\end{align*}

\vspace{-0.5em}
\subsection{Overfit Detection}
\vspace{-0.5em}
\label{ssec:overfit_detect}

The segmentation masks as shown in Fig.~\ref{fig:subset_aug_seg} and Fig.~\ref{fig:distortions} are constructed from attention maps and unique hues assigned to each class. The attention map is a grayscale image obtained by applying LSE on the CAM and then passing it through Sigmoid. For each pixel, it is closer to 1 if any class has been detected, and thus represents the model's decision region. 

The appearance of the attention maps is indicative of a model's degree of overfit. When a model overfits, it perceives even the most obscure features as being meaningful. Therefore, as can be seen in Fig.~\ref{fig:subset_aug_seg}, \emph{the attention area becomes smaller, less dilated, and has clearer boundaries as the number of training samples increases}, the segmentation masks also tend to identify the pixels within a single image as having features of only one class instead of multiple classes. This pattern exists regardless of the data augmentation applied to the model, and in fact regardless of the distortions present in the testing images as well. A simple quantitative measure is derived from the attention maps that offers a clear and intuitive expression of the model's overfitting behavior and can be used to determine the relative levels of overfit among multiple models. In the experiments performed, it exhibits an increasing trend for all data augmentation settings as the subset used to train the model gets larger, as demonstrated by Table \ref{table:sparsity}. This \emph{sparsity} measure calculates the percentage of zeros in each attention map averaged over the testing set:

\vspace{-1.5em}
\[ Sparsity=\frac{1}{I}\sum_{i=0}^{I-1} \frac{\sum_{m=0}^{M-1} \sum_{n=0}^{N-1} \mathds{1}[round(A_{i}[m,n])=0]}{MN} \]
where $I=$ total number of testing images, $A_{i}=$ attention map for image $i$, $M, N=$ dimensions of the attention maps (same as the image size), and $\mathds{1}$ is the indicator function.
\vspace{-0.5em}
\section{Experiment}
\label{sec:experiment}
\vspace{-0.5em}

\begin{table*}[t!]
\centering
\scalebox{0.73}{
\begin{tabular}{ccccccccccccccccccccc}
\toprule
& \multicolumn{2}{c}{\textbf{10\%}} & \multicolumn{2}{c}{\textbf{20\%}} & \multicolumn{2}{c}{\textbf{30\%}} & \multicolumn{2}{c}{\textbf{40\%}} & \multicolumn{2}{c}{\textbf{50\%}} & \multicolumn{2}{c}{\textbf{60\%}}& \multicolumn{2}{c}{\textbf{70\%}} & \multicolumn{2}{c}{\textbf{80\%}} & \multicolumn{2}{c}{\textbf{90\%}} & \multicolumn{2}{c}{\textbf{100\%}} \\
& I & II & I & II & I & II & I & II & I & II & I & II & I & II & I & II & I & II & I & II \\
\midrule[\heavyrulewidth]
\textbf{None} & 57.7 & 0.157 & 69.5 & 0.114 & 74.6 & 0.097 & 79.2 & 0.075 & 81.8 & 0.062 & 84.0 & 0.068 & 85.5 & 0.074 & 86.7 & 0.078 & 87.4 & 0.066 & 88.6 & 0.066 \\
\midrule
\textbf{Crop} & 65.8 & 0.209 & 77.2 & 0.102 & 83.3 & 0.065 & 86.6 & 0.045 & 88.5 & 0.036 & 90.0 & 0.032 & 91.6 & 0.024 & 92.2 & 0.020 & 92.8 & 0.018 & \textbf{93.5} & \textbf{0} \\
\midrule
\textbf{Hor. Flip} & 63.3 & 0.161 & 74.8 & 0.100 & 81.1 & 0.062 & 83.8 & 0.050 & 86.7 & 0.043 & 88.3 & 0.043 & 89.6 & 0.035 & 90.9 & 0.039 & 91.8 & 0.040 & 92.4 & 0.044 \\
\midrule
\textbf{Rotate} & 65.4 & 0.112 & 75.9 & 0.065 & 81.7 & 0.064 & 84.7 & 0.051 & 87.0 & 0.043 & 88.6 & 0.041 & 90.0 & 0.040 & 91.3 & 0.040 & 91.4 & 0.048 & 92.3 & 0.049\\
\midrule
\textbf{Persp.} & 64.4 & 0.184 & 76.5 & 0.094 & 82.6 & 0.067 & 85.5 & 0.062 & 88.0 & 0.051 & 89.6 & 0.045 & 90.7 & 0.044 & 91.7 & 0.046 & 92.2 & 0.031 & 93.2 & 0.032 \\
\midrule
\textbf{Color} & 64.4 & 0.125 & 72.0 & 0.097 & 76.8 & 0.081 & 79.4 & 0.070 & 82.0 & 0.061 & 84.7 & 0.064 & 85.7 & 0.059 & 87.0 & 0.055 & 87.1 & 0.062 & 88.5 & 0.070 \\
\midrule
\textbf{Comb.} & 55.5 & 0.295 & 69.8 & 0.187 & 77.6 & 0.136 & 82.7 & 0.104 & 85.9 & 0.084 & 88.0 & 0.070 & 89.4 & 0.061 & 90.2 & 0.050 & 91.8 & 0.040 & 92.2 & 0.037 \\
\bottomrule
\end{tabular}
}
\caption{Accuracy (\%) and MSE with respect to the best performing SCrIBE model (Crop 100\%). I $=$ Accuracy, II $=$ MSE.}
\label{table:acc_mse}
\vspace{-0.5em}
\end{table*}

\vspace{-0.5em}
\subsection{Setup}
\vspace{-0.5em}
\label{ssec:setup}

The dataset we use is CIFAR-10 \cite{cifar10}, which consists of 50,000 training images and 10,000 testing images of size $32 \text{px} \times 32 \text{px}$ with three color channels, divided into 10 classes. 
The model chosen is ResNet-18 \cite{resnet, pytorch}. The configuration comprises of separately training a regular classification model and a SCrIBE model with:
\begin{itemize}
  \item 10 subsets of training samples ranging from 10\% to 100\% of the training set;
  \item 7 data augmentation settings using transforms from the PyTorch library \cite{pytorch}: no transform, RandomCrop, RandomHorizontalFlip, RandomRotation, RandomPerspective, ColorJitter, and combining all of the above transforms;
\end{itemize}
and then testing both models on the original and perturbed images. 12 corruptions from ImageNet-C \cite{imagenet_c} are applied to the testing set, are shown in Fig.~\ref{fig:distortions}; each corruption ranges from severity 1 to 5.

Each model is trained for 30 epochs using a multi-step learning rate. 
The metrics we examine are accuracy, confidence, and expected calibration error (ECE) \cite{calib}. For the SCrIBE models, we also inspect the segmentation masks and attention maps.

\begin{table}[t!]
\centering
\scalebox{0.85}{
\begin{tabular}{ccccccc}
\toprule
& \multicolumn{2}{c}{\textbf{Acc. (\%)}} & \multicolumn{2}{c}{\textbf{Conf. (\%)}} & \multicolumn{2}{c}{\textbf{ECE (\%)}} \\
& I & II & I & II & I & II \\
\midrule[\heavyrulewidth]
\textbf{None} & 87.7 & \textbf{88.6} & 93.3 & \textbf{95.7} & \textbf{5.7} & 7.2 \\
\midrule
\textbf{Crop} & 92.4 & \textbf{93.5} & 95.1 & \textbf{97.1} & \textbf{2.7} & 3.6 \\
\midrule
\textbf{Hor. Flip} & 91.2 & \textbf{92.4} & 94.7 & \textbf{96.8} & \textbf{3.5} & 4.4 \\
\midrule
\textbf{Rotation} & 90.5 & \textbf{92.3} & 94.6 & \textbf{97.0} & \textbf{4.1} & 4.7 \\
\midrule
\textbf{Perspective} & 92.0 & \textbf{93.2} & 94.4 & \textbf{96.7} & \textbf{2.6} & 3.5 \\
\midrule
\textbf{ColorJitter} & 86.7 & \textbf{88.5} & 93.0 & \textbf{95.7} & \textbf{6.3} & 7.2 \\
\midrule
\textbf{Combined} & 90.0 & \textbf{92.2} & 90.6 & \textbf{94.1} & \textbf{0.7} & 2.2 \\
\bottomrule
\end{tabular}
}
\caption{Accuracy, confidence and expected calibration error of SCrIBE and non-SCrIBE models for each data augmentation. I $=$ non-SCrIBE, II $=$ SCrIBE. Models are trained with full training set.}
\label{table:scribe_non_scribe_cmp}
\vspace{-0.5em}
\end{table}

\vspace{-0.5em}
\subsection{SCrIBE and Non-SCrIBE Models Comparison}
\vspace{-0.5em}
\label{ssec:cmp}

As shown in Table \ref{table:scribe_non_scribe_cmp}, SCrIBE models have slightly higher accuracy, confidence, and ECE for all augmentations, but the differences are not significant. Additionally, they exhibit very similar performance and calibration trends for various training configurations and testing inputs. Therefore, the SCrIBE models can replace the non-SCrIBE models without much effect on performance, while providing visualizations that offer insight into the model's decision-making.

\vspace{-0.5em}
\subsection{Accuracy and Attention Region Alignment}
\vspace{-0.5em}
\label{ssec:acc_attn}

Although accuracy is a universal measure used in evaluating model performance, accuracy alone is not truly characteristic of a model's predictive power, since the correct decision might be made for the wrong reasons. Based on experimental results, models trained with RandomCrop and the full training set provide the highest accuracy when tested on uncorrupted images. We use this as the baseline, calculate the mean squared error (MSE) between each image's attention map acquired using models trained with other configurations and the attention map obtained through this optimal model, then average all MSE over the testing set. In this way, we can observe to what degree the model's decision areas match that of the optimal model, and whether its accuracy reflects this alignment. The various training configurations' accuracy and MSE are included in Table \ref{table:acc_mse}. We see that models that have high accuracy do not necessarily have a low MSE, and vice versa. This points to the importance of incorporating spatial domain visualizations in assessing model performance.

\begin{figure*}[h!]
  \centering
  \includegraphics[width=1\linewidth]{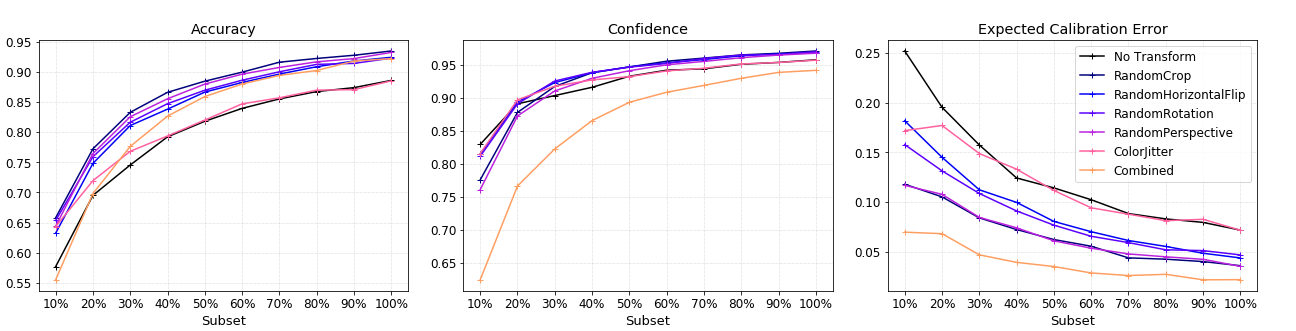}
  \caption{Performance and calibration plots for SCrIBE models when tested on uncorrupted images.}
  \label{fig:plots}
  \vspace{-0.5em}
\end{figure*}

\vspace{-0.5em}
\subsection{Data Augmentations}
\vspace{-0.5em}
\label{ssec:data_aug}


\emph{Data augmentations cause the attention regions to appear more smeared and less well-defined}. 
Since the goal of data augmentation is to make models more generalizable, data-augmented models avoid pinpointing sample-specific features and instead detect as large of an attention region as possible, \emph{causing the attention maps to be less sparse}, as shown in Table \ref{table:sparsity}. This behavior is more pronounced for smaller training sets and when multiple transforms are present. 

Fig.~\ref{fig:subset_aug_seg} illustrates that the attention regions are closely associated with the calibration of a model. The ordering for data augmentations based on decreasing number of disconnected clusters for small subsets (increasing attention area for large subsets) 
agrees with the ranking displayed in the ECE plot in Fig.~\ref{fig:plots} (from higher to lower error) across all subsets, demonstrating the level of calibration each transform provides. 
The above pattern is also displayed quantitatively in the sparsity measures in Table \ref{table:sparsity}. This behavior suggests that \emph{small attention areas are indicative of poorly calibrated models}.

\begin{figure*}[h!]
  \centering
  \includegraphics[width=0.75\linewidth]{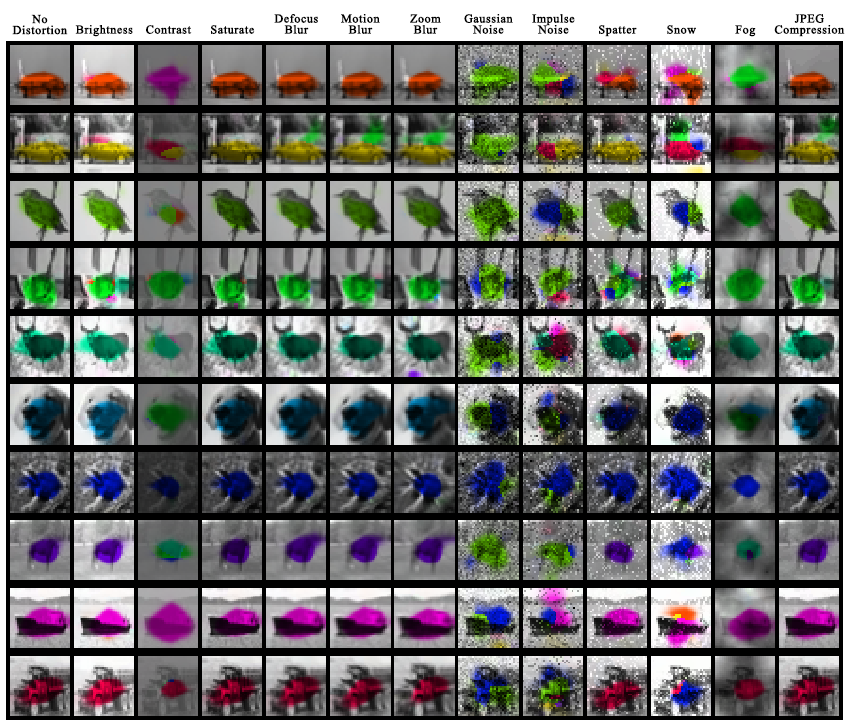}
  \caption{Sample corrupted segmentation masks (level 3 severity). Model used here is trained with full training set and RandomCrop.}
  \label{fig:distortions}
  \vspace{-0.5em}
\end{figure*}

\vspace{-0.5em}
\subsection{Input Perturbations}
\vspace{-0.5em}
\label{ssec:input_perturb}


All perturbations degrade performance, but brightness, saturate, blurring, and JPEG compression do not affect the performance drastically. On the other hand, corruptions related to noise cause the performance and calibration trends to be chaotic even at low levels.

\emph{Corruptions that remove details} from images tend to trick the model into \emph{detecting classes that depend on shapes and contours}; on the other hand, \emph{corruptions that add distracting information} cause the model to have \emph{false detection of classes that rely on texture}. For example, when images corrupted with high levels of brightness are tested on a highly overfit model, many images are incorrectly seen to have plane and ship features, possibly because increased brightness destroys details that are features of other classes. Similarly, when encountering images with decreased contrast, the models tend to detect deer, plane, and ship; these classes are recognized by features that have sharp corners and clean contours that stand out from the background. 
Gaussian noise, impulse noise, spatter, and snow tend to trick the model into falsely perceiving images as having frog and bird features, since the textures of frogs and birds imitate noise. This effect is especially prominent in Gaussian and impulse noise, where even low levels of noise significantly disrupt the segmentation masks, and for higher levels of noise, almost all images are seen to have frog and bird features. 
High levels of increased saturation exhibit similar behavior to noise distortions due to artifacts introduced by severe saturation that resemble noise. In the case of fog distortion, many images, and almost all images for the combined augmentation case, are detected to have cat features, possibly because the cloudy distortion imitates the fur of cats. 
Fig.~\ref{fig:distortions} provides an illustration of the segmentation masks behavior discussed above.

\emph{Data augmentations that are related to and compensate for the distortion generally exhibit the best performance and calibration}. For example, ColorJitter has a clear advantage over other augmentations when tested with brightness, contrast, saturation, and fog distortions. RandomPerspective is the most effect with defocus blur and motion blur, since manipulating the perspective might induce stretching and blurring that can be learned during training. RandomCrop works the best with zoom blur, because images corrupted with zoom blur are not only blurred but also slightly zoomed-in, causing some of the outer edges to be cut off. RandomRotation helps the most with Gaussian noise, impulse noise, spatter, and snow, often providing reasonable trends when results from other data augmentations appear chaotic, since it trains the model to look for high level contours. Combining all augmentations usually results in the lowest calibration error but never the highest accuracy for any distortion.
\vspace{-0.5em}
\section{Conclusion}
\label{sec:conclusion}
\vspace{-0.5em}
We have demonstrated that the segmentation masks obtained through a SCrIBE model provide intuitive insight on the spatial features the model inspects in making a decision under different scenarios such as the model's degree of overfitting, the data augmentation used, and the distortions present in the testing set. We have discussed the model's performance and calibration in each case and concluded that a SCrIBE model can be used in place of the original model without compromising performance. The relationship and trade-off among the configurations have been explored, and an overfit detection metric is derived from the attention maps without using testing labels.

\bibliographystyle{IEEEbib}
\bibliography{refs}

\end{document}